\pdfoutput=1

\documentclass[11pt]{article}

\usepackage[final]{acl}

\usepackage{times}
\usepackage{latexsym}

\usepackage[T1]{fontenc}

\usepackage[utf8]{inputenc}

\usepackage{microtype}

\usepackage{inconsolata}

\usepackage{graphicx}

\usepackage{booktabs}
\usepackage{multirow}
\usepackage{enumitem}
\usepackage{soul}
\usepackage{tabularray}
\usepackage{caption}
\usepackage{amsmath}
\usepackage{comment}
\usepackage{longtable}

\usepackage{newfloat}
\usepackage{listings}
\usepackage{tcolorbox}

\tcbuselibrary{skins,breakable}

\usepackage{colortbl}
\usepackage{amssymb}

\colorlet{transparentred}{red!30}
\colorlet{transparentgreen}{green!30}

\definecolor{myred}{rgb}{0.75, 0, 0}
\definecolor{mygreen}{RGB}{0, 150, 0}

\newcommand{\myindent}{\hspace*{1.55em}}
\newcommand{\model}{ChatGE}

\newcommand\blfootnote[1]{%
  \begingroup
  \renewcommand\thefootnote{}\footnote{#1}%
  \addtocounter{footnote}{-1}%
  \endgroup
}

\title{Game Development as Human-LLM Interaction}

\author{
    Jiale Hong$^*$,
    Hongqiu Wu$^*$,
    Hai Zhao$^\dag$ \\
    Department of Computer Science, Shanghai Jiao Tong University\\
    \{hongjiale, wuhongqiu\}@sjtu.edu.cn, zhaohai@cs.sjtu.edu.cn
}

\begin{document}

\maketitle

\blfootnote{$^*$ Equal contribution; $^\dag$ Corresponding author.}

\begin{abstract}

Game development is a highly specialized task that relies on a complex game engine powered by complex programming languages, preventing many gaming enthusiasts from handling it. This paper introduces the \textit{Chat Game Engine (\model{})} powered by LLM, which allows everyone to develop a custom game using natural language through Human-LLM interaction. To enable an LLM to function as a \model{}, we instruct it to perform the following processes in each turn: (1) $P_{script}$: configure the game script segment based on the user’s input; (2) $P_{code}$: generate the corresponding code snippet based on the game script segment; (3) $P_{utter}$: interact with the user, including guidance and feedback. We propose a data synthesis pipeline based on LLM to generate game script-code pairs and interactions from a few manually crafted seed data. We propose a three-stage progressive training strategy to transfer the dialogue-based LLM to our \model{} smoothly. We construct a \model{} for poker games as a case study and comprehensively evaluate it from two perspectives: interaction quality and code correctness.
\end{abstract}

%

\begin{figure}[ht]
  \centering
  \includegraphics[width=1\linewidth]{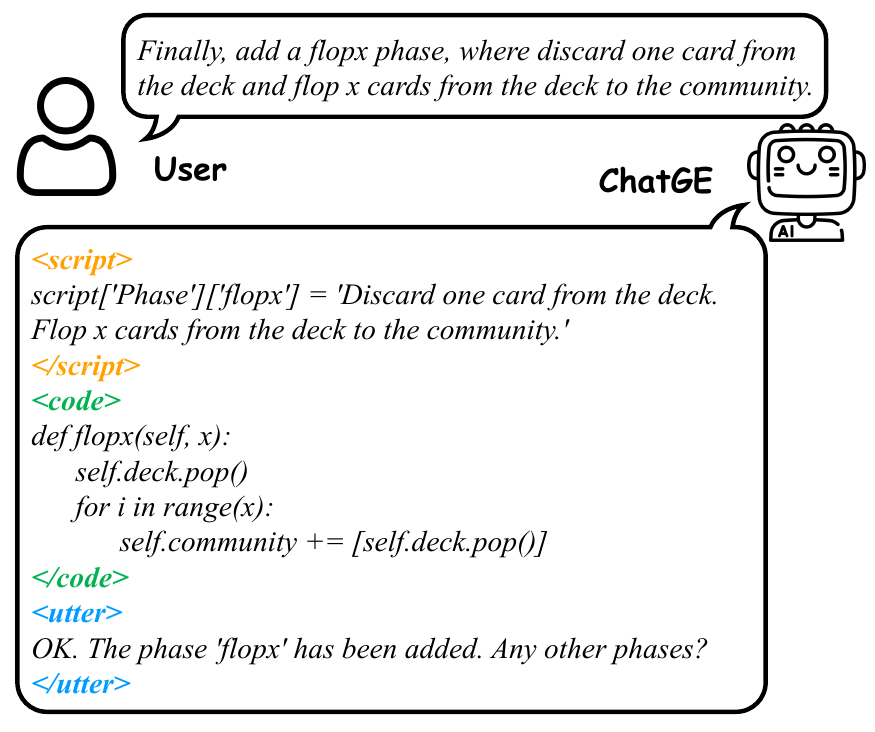}
  
  \caption{An illustration of the \model{} process for a poker game. See Appendix~\ref{app-sec:example} for a complete example.}
  \label{fig:example}
  
\end{figure}

\section{Introduction}
A game engine is a software framework primarily designed for the development of games and generally includes relevant libraries and support programs~\cite{valencia2016technologies}. Game engines help streamline the game development process, enabling developers to focus more on designing gameplay and content. Popular game engines include Unity, Unreal Engine, CryENGINE, etc.

Game development is a highly specialized task that relies on game engines powered by complex programming languages. 
The learning curve can be steep for those who wish to develop games based on their own designs. To make game development accessible to everyone, we propose the \textit{Chat Game Engine (\model{})}, powered by LLMs~\cite{brown2020language, achiam2023gpt, touvron2023llama}. 
This engine is designed to support the development of custom games using natural language through Human-LLM interaction. 

Compared to traditional game engines, our \model{} eliminates the learning curve. While traditional game engines provide users with software interfaces powered by complex technologies and programming languages, our \model{} offers a more flexible natural language interface powered by LLM. One can simply input natural language under the guidance of the engine through Human-LLM interaction. In \model{}, a user's natural language input is equivalent to calling software interfaces in a traditional game engine. The LLM generates implementation code based on the user's input, mirroring the process of implementing software interfaces through complex technologies and programming languages in traditional game engines.

\model{} is based on large language models (LLMs), which have shown exceptional capabilities in natural language processing across various aspects. In this work, we explore the joint capability of interaction and programming of the LLM to serve as a game engine, enabling development through natural language via Human-LLM interaction. As illustrated in Figure~\ref{fig:example}, we instruct the LLM to perform the following processes in each turn: \textbf{(1) $P_{script}$}: configure the game script segment based on the user’s input; \textbf{(2)$P_{code}$}: generate the corresponding code snippet based on the game script segment; \textbf{(3) $P_{utter}$}: interact with the user, including guidance and feedback.

We propose a comprehensive training paradigm to fine-tune an LLM to excel as a \model{}, rather than relying solely on prompting. 
There are two main challenges. 
First, it is an exhausting process to acquire a large number of game script-code pairs. We propose an efficient data synthesis pipeline to generate game script-code pairs automatically from a few manually crafted seed data. 
Moreover, our framework requires the LLM to perform $P_{script}$, $P_{code}$, and $P_{utter}$ step by step, which challenges the joint capability of interaction and programming. Additionally, a straightforward strategy to train on sufficient complete interaction data is inefficient.
Therefore, we propose a three-stage progressive training strategy to transfer the dialogue-based LLM to our \model{} smoothly.

Eventually, we construct a \model{} for Poker, a worldwide card game, e.g. \textit{Texas hold’em}. We utilize the proposed data synthesis pipeline to generate the corresponding dataset and fine-tune a \model{} using the presented strategy. Then we propose a fine-grained evaluation process, measuring the performance from two perspectives: interaction quality and code correctness.

In summary, this paper:
\begin{itemize}
\item introduces the \model{} framework for game development as Human-LLM interaction;
\item presents the data generation technique that fuels the learning of \model{};
\item proposes a three-stage progressive training strategy for effectively training \model{};
\item constructs a \model{} for poker games and evaluates its performance from two perspectives: interaction quality and code correctness.
\end{itemize}

\section{Related works}

\paragraph{AI for Games}
AI for games is an exciting area in AI research. A great amount of recent work studies learning for agents, e.g. as game players for Atari ~\citep{DBLP:journals/corr/MnihKSGAWR13}, Minecraft ~\citep{DBLP:conf/nips/FanWJMYZTHZA22,DBLP:journals/corr/abs-2305-16291}, StarCraft, ~\citep{DBLP:journals/nature/VinyalsBCMDCCPE19}, NetHack ~\citep{DBLP:conf/nips/KuttlerNMRSGR20,DBLP:conf/iclr/Lowe0FKP20}, Werewolf ~\citep{DBLP:journals/corr/abs-2309-04658}; non-play characters (NPCs) ~\citep{DBLP:journals/nature/ShanahanMR23,DBLP:journals/air/UludagliO23}; player assistants ~\citep{gallotta2024large}; game commentators ~\citep{DBLP:conf/icids/Eladhari18,DBLP:conf/exag/RanellaE23}.
Recently, there has been work focused on building a neural engine based on LLMs.
IDGE~\cite{wu2024instruction} autoregressively predicts in-game states based on player actions, functioning more like a game runtime environment that supports game creation by simple natural language instructions as a script.
In comparison, our \model{} serves as a development framework for creating games, similar to a traditional game engine.

\paragraph{LLMs as Training Data Generators}
With the immense power demonstrated by large language models(LLMs), researchers have recently explored their potential as as training data generators~\cite{yu2024large}. Such applications include generating tabular data~\cite{borisov2022language}, medical dialogue~\cite{chintagunta2021medically}, sentence pairs~\cite{schick2021generating}, role-play dialogue~\cite{shao2023character}, instruction data~\cite{peng2023instruction, shao2023synthetic, sun2024principle, wang2022self}, etc.. In this paper, we propose a data synthesis pipeline that leverages LLMs as training data generators to produce game script-code pairs and user-LLM interactions from a few manually crafted seed data.

\begin{figure*}[ht]
  \centering
  \includegraphics[width=1\textwidth]{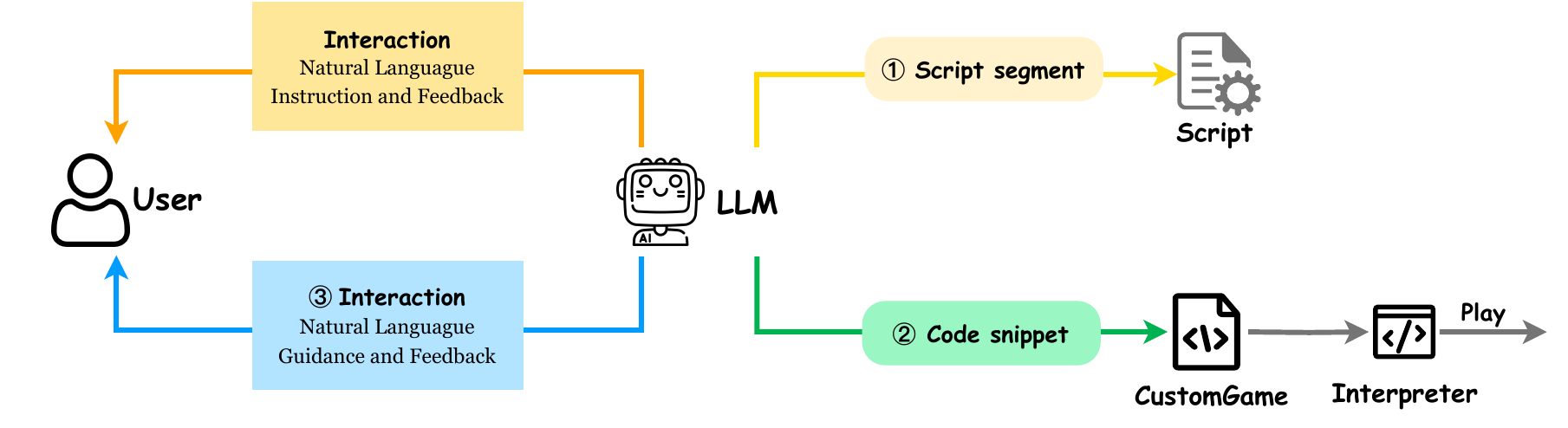}
  
  \caption{\model{} framework. The LLM processes the user's input in the orange stream, while simultaneously generating script in the yellow stream, code in the green stream, and interaction in the blue stream.}
  \label{fig:framework}
  
\end{figure*}

\paragraph{Progressive Training Strategy}
Progressive training strategy is commonly employed in LLM training. Training on progressively increasing sequence length data in multi-stages is used to mitigate computational costs and enhance data efficiency in both the pre-training~\cite{jin2023growlength, dubey2024llama} and post-training~\cite{liu2024world} phases.
Curriculum learning ~\cite{bengio2009curriculum}, a specialized form of progressive training, gradually increases the complexity of data samples during the training process.
Recent studies show the promising role of curriculum learning in empowering the language models to tackle more challenging tasks~\cite{vakil2023complexity,wu2023empower,wu2024instruction}.
In this paper, we propose a three-stage progressive training strategy to transfer the dialogue-based LLM to our \model{} smoothly. This strategy also aligns with the principles of curriculum learning.

\section{\model{}}

In this section, we present our \model{} framework, illustrated in Figure~\ref{fig:framework}. 

  
  

\subsection{Overview}


The \model{} framework introduces a new paradigm of game development as Human-LLM interaction. 
In user-LLM interactions, the user provides instructions for their game concept in natural language under the guidance of LLM, along with feedback to the LLM. The LLM guides the user in refining and clarifying essential details about the game, while also offering feedback. To enable the LLM to provide effective guidance, we predefine a generic script template tailored to a specific type of game. 
Except for interaction with the user, the LLM generates script segments and code snippets to implement the user's game concept in each turn. Meanwhile, the code snippets are stored, building toward the eventual complete game code, \textit{CustomGame}. After the game is fully developed through multi-turn interactions, a code interpreter is used to execute the \textit{CustomGame} code for play.

\subsection{Formulation}
The complete process of \model{} framework can be seen as a multi-turn human-LLM interaction. We first formulate the multi-turn Human-LLM interaction and then extend this concept to \model{}. 

In a multi-turn Human-LLM interaction, both the user input and the LLM's output may be related to the interaction history, such as references to prior content. The interaction history $h_t$ at turn $t$ can be simply defined as:
\begin{equation}
\begin{aligned}
h_{t} = 
\begin{cases} 
\emptyset & \text{if } t = 0 \\
\left\{ (i_\tau, o_\tau) \mid \tau = 1, 2, \ldots, t \right\} & \text{if } t > 0 
\end{cases}
\end{aligned}
\end{equation}
where the subscript $t$ refers to the increasing number of turns, $i_t$ refers to the user input and $o_t$ refers to the LLM's output, formulated as:
\begin{equation}
\begin{aligned}
o_t &= \mathcal F_\theta(h_{t-1}, i_t)
\end{aligned}
\end{equation}
where $\mathcal F_\theta$ refers to the LLM, and $\theta$ denotes its parameters. Consequently, an LLM with parameters $\theta$ seeks to maximize the likelihood: $\sum_{t=1}^{T}\log p_\theta(o_t|h_{t-1}, i_t)$, where $T$ refers to the total number of interaction turns.

\begin{figure*}[ht]
  \centering
  \includegraphics[width=1\textwidth]{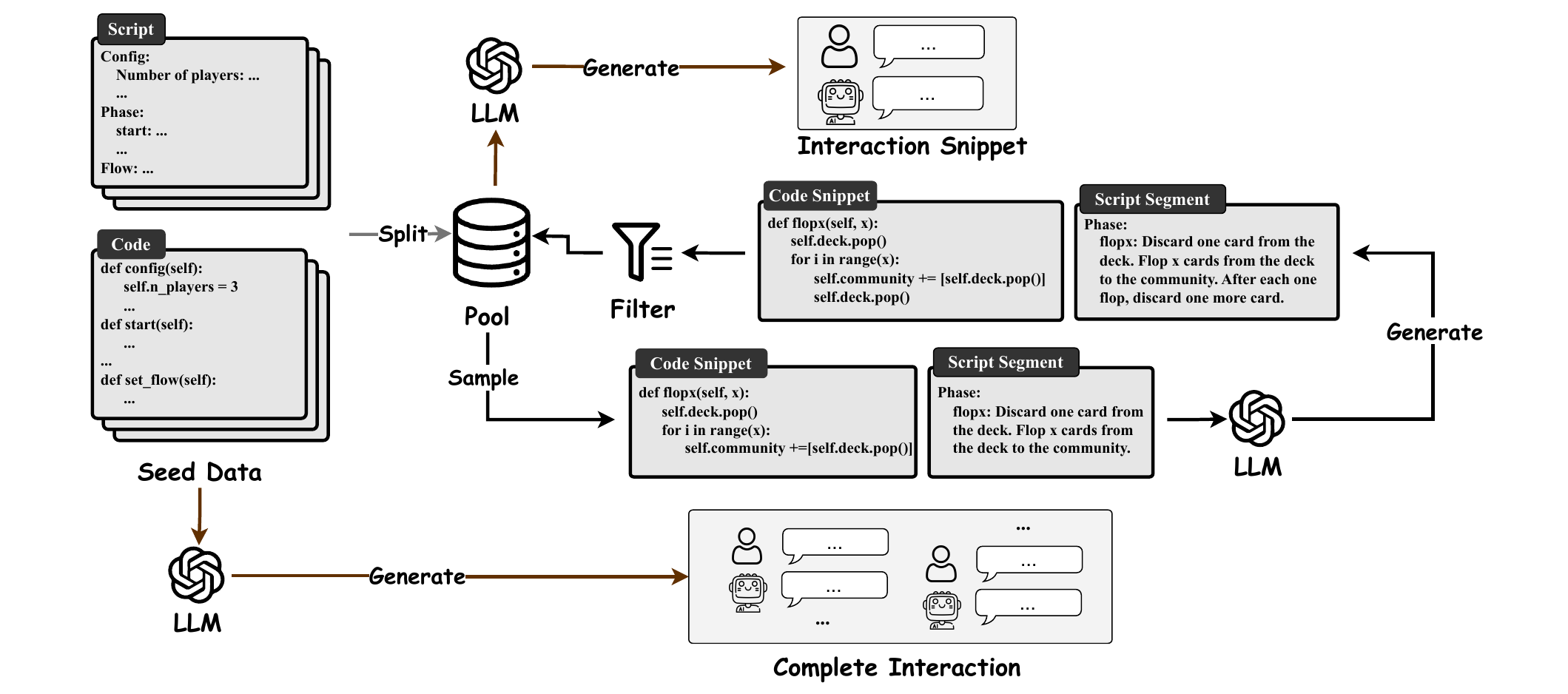}
  
  \caption{Data synthesis pipeline for game script-code pair and interaction generation.}
  \label{fig:data_generation}
  
\end{figure*}

The distinction between \model{} and a general multi-turn Human-LLM interaction lies in the specialization of the input and output. The user input $i_t$ consists of instructions about their game concept and feedback to the LLM. The LLM's output $o_t$ includes both interactions with the user and code snippets to implement the user’s game concept in one turn. 
To enable an LLM to function as a \model{}, we instruct the LLM to perform the following processes in each turn: \textbf{(1)} $P_{script}$: configure the game script segment based on the user's input(Enclosed by \texttt{<script></script>}: in Figure~\ref{fig:example}); \textbf{(2)} $P_{code}$: generate the corresponding code snippet based on the game script segment(Enclosed by \texttt{<code></code>}: in Figure~\ref{fig:example}); \textbf{(3)} $P_{utter}$: interact with the user, including guidance and feedback(Enclosed by \texttt{<utter></utter>}: in Figure~\ref{fig:example}). 
For interaction and coding requirements, $P_{code}$ and $P_{utter}$ are essential. $P_{script}$ serves as an intermediate process, akin to the reasoning in chain-of-thought (CoT) ~\cite{wei2022chain}. Additionally, it can also act as a visual representation of the current development progress. Compared to code, a script is much easier for people to understand, especially those without a programming background. Therefore, $o_t$ can be specilized as:
\begin{equation}
\begin{aligned}
o_t = (s_t, c_t, u_t) = \mathcal F_\theta(s_t, c_t, u_t|h_{t-1}, i_t).
\end{aligned}
\end{equation}
where $s_t$, $c_t$, $u_t$ refer to the outputs of $P_{script}$, $P_{code}$, and $P_{utter}$ respectively.
Furthermore, the ultimate objective of this task, \textit{CustomGame} $C$ can be obtained by merging $c_t$ across all turns:
\begin{equation}
\begin{aligned}
C = Merge(c_1, c_2, \ldots, c_T)
\end{aligned}
\end{equation}
where $Merge$ denotes the merge function. 
Specifically, $m$ can be determined by the specific game implementation. In our inplementaion, we embed $c_t$ into the base code of the specific game.


\section{Data Generation} \label{sec:data_generation}

In this section, we discuss our attempt in data generation. Utilizing LLMs to create \model{} requires fine-tuning on a substantial amount of supervised data. However, manually crafting diverse interactions with script-code pairs is a challenging task. 
Compared to fully manual annotation, harnessing LLMs to synthesize data is more efficient and has become a popular method for addressing the issue of insufficient data. 
We propose a pipeline consisting of three main steps to synthesize data, starting with a small set of manually annotated seed data, as illustrated in Figure~\ref{fig:data_generation}.
We utilize GPT-4o as the generator.

\paragraph{Init pool} First, we manually craft a few script-code pairs, each corresponding to different custom games. These pairs serve as seed data and are then split into script segments and code snippets, which are added to the pool.

\paragraph{Generate new pairs} 
In this step, we sample pairs of script segments and code snippets, generating new pairs based on these selections.
We prompt the generator to modify the code snippet first, then generate the corresponding script segment. This order is chosen because it’s easier to map a script segment to a code snippet by describing the code, rather than generating code from a script description. This approach yields higher-quality data with more reliable mappings. To ensure the generated code functions correctly, we implement a filter to discard code that fails to execute. This process continues until the pool contains a sufficient number of entries.

\paragraph{Generate interaction data} 
Finally, we generate the interaction data depicted in Figure~\ref{fig:example} using the script-code pairs. This process involves two lines: (1) generating interaction snippets based on pairs of script segments and code snippets from the pool; (2) generating complete interactions from complete script-code pairs. The necessity of these two data components will be discussed in the next section.

\section{Training Strategy}
\label{sec:train}
In this section, we present our training strategy. 
In our framework, the LLM will perform $P_{script}$, $P_{code}$ and $P_{utter}$ step by step, which challenges the joint capability of interaction and programming. On the other hand, a straightforward strategy to train on sufficient complete interaction data is inefficient.
Therefore, we propose a three-stage progressive training strategy to transfer the dialogue-based LLM to our \model{} smoothly. 

\paragraph{Stage-1: Base Training} This stage aims to train the base interaction ability of the model. Interaction ability is the most fundamental ability for \model{} and serves as the foundation for the following two stages. Since most LLMs have already undergone sufficient and efficient supervised fine-tuning (SFT)~\cite{brown2020language,raffel2020exploring,ouyang2022training}, we can directly use such models for Stage-1.

\paragraph{Stage-2: Core Training} This stage aims to train the core capabilities of the model, namely the joint capability of programming and interaction. It fine-tunes the model from Stage-1 on interaction snippets that follow the \model{} format. As illustrated in Figure~\ref{fig:example}, we instruct the model to perform the $P_{script}$, $P_{code}$ and $P_{utter}$ step by step to extract the user's concept of the game, implement it in code, and provide guidance and feedback for interaction.

\paragraph{Stage-3: Alignment} This stage aims to align the model with a complete interaction context to fully develop a game as a \model{}. It fine-tunes the model from Stage-2, which already possesses significant programming and interaction capabilities. At this stage, we only need to extend its ability for multi-turn interactions as a \model{}, particularly in guiding users to complete game development according to the predefined script. Since the model already possesses strong multi-turn interaction and long-context capabilities following Stage-1 training, only a small dataset is required for alignment at this stage.

\section{Experiments}\label{sec:experiments}

In this section, we construct a \model{} for a poker game. 
We employ the proposed data synthesis pipeline to generate the corresponding dataset, fine-tune a \model{} using the presented strategy and evaluate its performance.

\subsection{Dataset}


\begin{table}[ht]
\renewcommand\arraystretch{1}
\scriptsize
\resizebox{\linewidth}{!}{
\begin{tcolorbox}[colback=white!95!gray,colframe=gray!50!black,rounded corners,label={box:rubrics},left=1mm, right=1mm, top=1mm, bottom=1mm, title={Texas hold'em}]
\textit{\textbf{Config}}: \\ 
\textbf{Number of players}: 3 \\ 
\textbf{Min bet}: 10 \\ 
\textbf{Max bet}: 1000 \\ 
\textbf{Suit}: H, D, C, S \\
\textbf{Suit have rank}: False \\ 
\textbf{Card value rank}: 2, 3, 4, 5, 6, 7, 8, 9, 10, 11, 12, 1\\ 
\textbf{Card combinations rank}: High Card, Pair, Two Pair, Three of a Kind, Straight, Flush, Full House, Four of a Kind, Straight Flush \\ 
\textit{\textbf{Phase}}: \\ 
\textbf{start}: Config the game, prepare the deck and chips for all players. \\ 
\textbf{blind}: Randomly assign two: small blind bets minimum, big blind double. \\ 
\textbf{dealx}: Deal x cards to each player. \\ 
\textbf{switch}: Each discards and draws the same number of cards from the deck. \\ 
\textbf{bet}: Each bets until all the unfolded match the highest or only 1 remains. \\
\textbf{flopx}: Discard a card from the deck. Reveal x community cards. \\ 
\textit{\textbf{Flow}}: start, blind, deal2, bet, flop3, bet, flop1, bet, flop1, bet, show, prize
\end{tcolorbox}
}
\caption{An example game script for a poker game.}
\label{tab:game_script}
\end{table}

\paragraph{Poker Game}
Poker, a worldwide card game, \textit{e.g. Texas hold’em, Badugi}. These poker games can be abstracted into a generic game script. Table~\ref{tab:game_script} presents an example example of such a script for the classic \textit{Texas hold’em}. This generic script allows for the configuration of several common elements across different poker games, including the number of players, minimum and maximum bet limits, suit types and rankings, single-card rankings, multi-card combination rankings, game phases, and overall game flow. By adjusting these elements, virtually infinite variations of poker can be created. Notably, each game in our dataset corresponds to a unique configuration, including customizable phases. For example, a standard ``flopx'' phase might involve discarding one card from the deck and then revealing x community cards. This phase can be customized by adding a rule such as, ``After each flop, discard one more card,'' thereby creating a new variant of the ``flopx'' phase.

\paragraph{Data Statistics}

\begin{table}[!ht]
    \centering
    \resizebox{\linewidth}{!}{
    \setlength{\tabcolsep}{12pt}
    \begin{tabular}{l c c }
    \toprule
         \textbf{Statistics} & Training & Test \\ \hline
         \rowcolor{gray!20} \textit{manually crafted(seed data for training)} & & \\
         \# of games & 20 & 10 \\
        \# of script-code pairs(functions) & 180 & 90\\
        \midrule
        \rowcolor{gray!20} \textit{synthesis} & &\\
        \# of complete interactions & 36 & / \\
        \# of interaction snippets & 3718 & / \\
    \bottomrule
    \end{tabular}}
    \caption{Statistics of training and test data.}
    \label{tab:statistics}
\end{table}

Table~\ref{tab:statistics} shows the statistics of the training and test data that we construct. The interaction data format follows Figure~\ref{fig:example}. 

\begin{table*}[ht]
\resizebox{\textwidth}{!}{
    \renewcommand\arraystretch{1}
    \centering
    \scriptsize 
    \begin{tabular}{p{0.055\textwidth}p{0.58\textwidth}|p{0.09\textwidth}p{0.3\textwidth}}
    \toprule
    \multicolumn{2}{c|}{{\textbf{Evaluation Metrics}}} & \multicolumn{2}{c}{\textbf{Scoring Guide}} \\
    \textbf{Metric} & \textbf{Description} & \textbf{Score} & \textbf{Criteria} \\
    \midrule
    Guidance  & \scriptsize{How the response guide the user step-by-step to complete the game.} & 1~Poor & \scriptsize{Significant deficiencies or inaccuracies.} \\
    Logic & \scriptsize{Logical structure and soundness of reasoning, including the support and validity of conclusions.} & 2~Below Avg.   & \scriptsize{Noticeable weaknesses, lacking in several areas.}  \\
    Relevance & \scriptsize{The extent to which the response stays on topic and within the scope of the assistant role.}  & 3~Above Avg.   & \scriptsize{Mostly on target with a few minor shortcomings.}   \\
    Coherence & \scriptsize{Integration into the context, consistency with previous statements and conversational flow.} & 4~Strong  & \scriptsize{Strong performance, often surpasses expectations.} \\
    Conciseness  & \scriptsize{Brevity and clarity of the response, avoiding unnecessary elaboration or repetition.} &  & \\
    \bottomrule
    \end{tabular}
}
\caption{Evaluation Metrics and Scoring Guide. We design the criteria following ~\citet{yu2024kieval,wu2024role,zheng2024judging,wang2023pandalm,guo2023evaluating}. }
\label{tab:interaction_metric}

\end{table*}

\subsection{Setup}
We employ LLaMA3.1-8B-Instruct\footnote{https://huggingface.co/meta-llama/Meta-Llama-3.1-8B-Instruct}~\cite{dubey2024llama} for Stage-1 and finetune it using LoRA ~\cite{hu2021lora} with $r = 8$, $\alpha = 32$, and a learning rate of 3e-4. We train 3 epochs on the 3718 interaction snippets for Stage-2 and 5 epochs on the 36 complete interactions for Stage-3.

To assess the performance of the LLM in a dynamic multi-turn interaction environment, we require a user to interact with the LLM, as demonstrated in our \model{} framework. 
Simulating the user using a rule-based approach is complex, and employing human annotators poses challenges related to inconsistent standards and high costs. 
To address these issues, we use GPT-4o-mini as the interactor to simulate the user, a practice increasingly adopted in dynamic multi-turn interaction environments~\cite{wang2023mint, terragni2023context, davidson2023user, sekulic2024reliable, luo2024duetsim, xiang2024simuser, yu2024kieval}.
For evaluation, we provide the interactor with a manually crafted game script and instruct them to treat it as the game concept they have in mind.
The interactor then interacts with the LLM, resulting in a multi-turn interaction about a specific custom game. This allows us to use the game script and its corresponding code as the ground truth for evaluating the generated interaction.

\subsection{Metrics}
We assess model performance from two perspectives: interaction quality and code correctness.

\begin{table*}[ht]
  \centering
  \resizebox{\textwidth}{!}{
  \renewcommand\arraystretch{1}
  \begin{tabular}{l|cccccc|cccc}
    \toprule[1.0pt]
    \multirow{2}{*}{\textbf{Model}} & \multicolumn{6}{c|}{\textbf{Interaction Quality}} & \multicolumn{4}{c}{\textbf{Code Correctness}} \\
    & {Gui.} & {Log.} & {Rel.} & {Coh.} & {Con.} & {Overall} & {F-ESR} & {F-Acc} & {ESR} & {Acc} \\
    \midrule[0.6pt]
    \rowcolor{gray!20} \multicolumn{11}{c}{\textit{5-shot}} \\
    GPT-3.5-turbo & 94.5 & 96.5 & 100 & 99.0 & 96.5 & 98.0 & 95.8 & 87.9 & 60.0 & 30.0 \\
    GPT-4o & 98.5 & 98.0 & 100 & \textbf{100} & 99.0 & 99.0 & 93.0 & 88.5 & 50.0 & 30.0\\
    Llama-3.1-8B-Instruct & 97.5 & 98.0 & 100 & 99.5 & 99.0 & 98.5 & 100 & 90.0 & 60.0 & 10.0 \\
    \midrule[0.6pt]
    \rowcolor{gray!20} \multicolumn{11}{c}{\textit{Fine-tuning}} \\
    \model{} & \textbf{98.5} & \textbf{99.0} & \textbf{100} & 99.5 & \textbf{99.0} & \textbf{100} & \textbf{100} & \textbf{99.0} & \textbf{100} & \textbf{90.0} \\
    \quad \textit{w/o. $P_{script}$} & 98.0 & 97.0 & 100 & 99.0 & 96.5 & 98.5 & 100 & 98.8 & 100 & 80.0 \\
    \quad \textit{w/o. synthesis} & 96.5 & 96.0 & 100 & 98.0 & 96.5 & 98.0 & 97.4 & 86.8 & 70.0 & 0 \\
    \quad \textit{w/o. Stage-2} & 96.5 & 97.5 & 100 & 99.0 & 96.0 & 98.5 & 98.2 & 89.2 & 80.0 & 10.0 \\
    \quad \textit{w. Mixed-stage} & 92.5 & 96.5 & 99.5 & 96.0 & 92.0 & 95.0 & 95.0 & 88.5 & 80.0 & 20.0 \\
    \bottomrule[1.0pt]
  \end{tabular}
  }
  \caption{\label{tab:main_results} Main results of different models and the ablation study of \model{}. 
  The number of functions generated by the model can vary due to factors such as repeated modifications or missed queries.
  Functional-level metrics primarily assess the correctness of the generated code without accounting for recall rate, which is instead reflected in the overall-level metrics.}
\end{table*}

\paragraph{Interaction Quality} 
The interaction quality is assessed by an evaluator model, which assesses the output for guidance, logic, relevance, coherence and conciseness. Following KIEval~\cite{yu2024kieval}, we implement a scoring system to quantitatively grade model performance in different aspects. Responses are rated on a definitive scale from 1 to 4 for each aspect, where 1 and 4 denote `Poor’ and `Strong’ performance, respectively, as detailed in Table~\ref{tab:interaction_metric}. These scores are designed to encourage decisive evaluations. To facilitate comparison, we normalize the scores, ensuring that a rating of 1.0 indicates perfect performance. We utilize GPT-4o as the evaluator, run 5 times and average the results.

\paragraph{Code correctness}
We evaluate code correctness using two functional-level metrics and two overall-level metrics to respectively measure correctness of functions as code snippet and complete code:
\begin{itemize}[leftmargin=10pt]
\item \textbf{\textit{F-ESR}} represents the \textit{functional execution success rate} across the entire test set to measure the model's basic coding capability.
\item \textbf{\textit{F-Acc}} represents the \textit{functional accuracy} of the code, assessed through black-box testing to determine if the generated code is correct. Specifically, we replace player input with random input and, for each run, fix the random seed. We then compare the resulting game states after multiple turns until the game ends of the generated code with the ground truth. We conduct 40 runs, each with a different random seed, for every entry. If all runs produce identical states, the code is considered correct.

\item \textbf{\textit{ESR}} represents the \textit{execution success rate} of the complete code of a custom game.

\item \textbf{\textit{Acc}} represents the \textit{accuracy} of the complete code of a custom game. 
\end{itemize}

\subsection{Main Results}
We evaluate \model{} on the test data we construct, as shown in Table~\ref{tab:statistics}.
Table~\ref{tab:main_results} presents the performance of our \model{}, including both interaction quality and code correctness. 
For comparison, we take several representative closed-source and open-source LLMs in a 5-shot setting as baselines.
Intuitively, \model{} excels in both interaction quality and code correctness.

\subsubsection{Interaction Quality}
All models exhibit high interaction quality.
Our \model{} excels across all dimensions, showcasing exceptional capabilities in interacting with the user throughout the interactive development process. 
Compared to Llama-3.1-8B-Instruct, our fine-tuned model excels in guidance and logic, effectively guiding the user to develop the game logically.

\subsubsection{Code Correctness}
In our results, all models significantly outperform in functional-level metrics compared to overall-level metrics. 
This suggests that while LLMs excel at producing functional code, they face challenges when generating long, complete code. 
Additionally, it is evident that executability is more easily achieved than accuracy across all models, with our model reaching a perfect ESR of 100. This indicates that LLMs excel at generating code that is syntactically executable.
Notably, \model{} outperforms in all metrics. It achieves an impressive F-Acc of 99.0, outperforming the second-best model by 9 points. Moreover, it reaches an ESR of 100, surpassing the second-best by 20 points. Furthermore, it attains an Acc of an astounding 90, outstripping the second-best by 60 points. 

To conduct a more in-depth analysis, we compute the function-level code correctness in Table~\ref{tab:function_results}. 
Most models excel on fixed functions and two simple variable functions: config and flow. These two functions require only basic assignment statements to configure the game, allowing them to generalize effectively. 
However, for functions with more complex code logic, namely blind, dealx, and flopx, the baselines generally underperform, with the lowest F-Acc reaching just 20.
These results indicate that the accumulation of errors across these functions leads non-fine-tuned models to exhibit low correctness in overall-level evaluation.
It is important to note that the model is required to be all-round at each function; otherwise, the overall performance will degenerate in a way of Buckets effect~\cite{wu2024instruction}. 
Delightfully, our \model{} achieves near-perfect performance across all functions, resulting in an Acc far exceeding the baselines.

\begin{table*}[ht]
  \setlength\tabcolsep{2.3pt} 
  \renewcommand\arraystretch{1.02}
  \huge
  \centering
  \resizebox{\textwidth}{!}{
  \begin{tabular}{l|cc|cc|cc|cc|cc|cc|cc|cc|cc}
    \toprule[1.0pt]
    \multirow{2}{*}{\textbf{Model}} & \multicolumn{2}{c|} {\textbf{config\textsuperscript{*}}} & \multicolumn{2}{c|} {\textbf{start}} & \multicolumn{2}{c|} {\textbf{blind\textsuperscript{*}}} & \multicolumn{2}{c|} {\textbf{dealx\textsuperscript{*}}} & \multicolumn{2}{c|} {\textbf{flopx\textsuperscript{*}}} & \multicolumn{2}{c|} {\textbf{switch}} & \multicolumn{2}{c|} {\textbf{bet}} & \multicolumn{2}{c|} {\textbf{flow\textsuperscript{*}}} & \multicolumn{2}{c} {\textbf{Overall}} \\
    & {F-ESR} & {F-Acc} & {F-ESR} & {F-Acc} & {F-ESR} & {F-Acc} & {F-ESR} & {F-Acc}  & {F-ESR} & {F-Acc}  & {F-ESR} & {F-Acc} & {F-ESR} & {F-Acc} & {F-ESR} & {F-Acc} & {F-ESR} & {F-Acc} \\
    \midrule[0.6pt]
    \rowcolor{gray!20} \multicolumn{19}{c}{\textit{5-shot}} \\
    GPT-3.5-turbo & 100 & 88.9 & 100 & 100 & 87.5 & 87.5 & 88.9 & 55.6 & 87.5 & 75.0 & 100 & 100 & 100 & 100 & 88.9 & 66.7 & 95.8 & 87.9 \\
    GPT-4o & 100 & 100 & 100 & 100 & 70.0 & 70.0 & 90.0 & 90.0 & 100 & 60.0 & 100 & 100 & 80.0 & 80.0 & 100 & 100 & 93.0 & 88.5 \\
    Llama-3.1-8B-Instruct & 100 & 100 & 100 & 100 & 100 & 100 & 100 & 50.0 & 100 & 60.0 & 100 & 100 & 100 & 100 & 100 & 66.7 & 100 & 90.0\\
    \midrule[0.6pt]
    \rowcolor{gray!20} \multicolumn{19}{c}{\textit{Fine-tuning}} \\
    \model{} & \textbf{100} & \textbf{100} & \textbf{100} & \textbf{100} & \textbf{100} & \textbf{100} & \textbf{100} & \textbf{90.0} & \textbf{100} & \textbf{100} & \textbf{100} & \textbf{100} & \textbf{100} & \textbf{100} & \textbf{100} & \textbf{100} & \textbf{100} & \textbf{99.0} \\
    \quad \textit{w/o. $P_{script}$} & 100 & 100 & 100 & 100 & 100 & 100 & 100 & 90.0 & 100 & 90.0 & 100 & 100 & 100 & 100 & 100 & 100 & 100 & 98.8 \\
    \quad \textit{w/o. synthesis} & 100 & 100 & 100 & 100 & 100 & 100 & 88.9 & 22.2 & 100 & 55.6 & 100 & 100 & 100 & 100 & 100 & 100 & 93.8 & 86.8 \\
    \quad \textit{w/o. Stage-2} & 100 & 100 & 100 & 100 & 100 & 100 & 90.0 & 20.0 & 90.0 & 60.0 & 100 & 100 & 100 & 100 & 100 & 100 & 98.2 & 89.2 \\
    \quad \textit{w. Mixed-stage} & 100 & 100 & 100 & 100 & 81.8 & 81.8 & 100 & 33.3 & 90.0 & 90.0 & 90.0 & 90.0 & 81.8 & 81.8 & 100 & 75.0 & 95.0 & 88.5\\
    \bottomrule[1.0pt]
  \end{tabular}
  }
  \caption{\label{tab:function_results} Function-level code correctness of different models and the ablation study of \model{}. Functions with an asterisk (*) are variable functions in the test set, while the remaining functions are fixed.
  }
  
\end{table*}

\subsection{Ablation Study}

We ablate different variants from the full \model{} architecture, the results are presented in Table~\ref{tab:main_results} and Table~\ref{tab:function_results}. Traning data statistics of ablations can be found in Appendix~\ref{app-sec:ablation-statistics}.

\paragraph{Ablation on $P_{script}$}
A slight decrease can be observed in interaction quality across nearly all dimensions without $P_{script}$. Additionally, F-Acc drops by 0.2 points and Acc by 10.0 points. As shown in Table~\ref{tab:function_results}, the only failure occurs on a flopx function when compared to the complete \model{} architecture. This suggests that $P_{script}$ can enhance both interaction and coding abilities in certain cases.

\paragraph{Ablation on synthetic data}
In this setting, we directly employ manually crafted script-code pairs, splitting them into snippets to generate complete interactions and interaction snippets. 
A slight decline can be observed in interaction quality across most dimensions, alongside a significant decrease in code correctness, with Acc dropping to 0. Notably, the code correctness is even lower than that of the 5-shot Llama-3.1-8B-Instruct. As shown in Table~\ref{tab:function_results}, this decline is attributed to poor performance on the two most challenging functions, dealx and flopx. 
This can be explained by the model overfitting on the limited data due to the absence of synthetic data, which leads to poor generalization.

\paragraph{Ablation on training strategy}
We conducted comprehensive ablation experiments on our three-stage training strategy, with the following setups: w/o. Stage-1, w/o. Stage-2, w/o. Stage-3, and w. Mixed-stage. 
In the setups without Stage-1 and Stage-3, the model loses its guiding and interaction abilities in multi-turn scenarios as a \model{}, resulting in ESR and Acc values of 0. 
Therefore, the results of these two settings are not reported. 
This suggests that both Stage-1 and Stage-3 play a crucial role in enhancing the model's interaction ability as a \model{}. As shown in Table~\ref{tab:main_results}, the interaction quality of the model decreases across most dimensions without Stage-2. 
Additionally, there is a significant drop in code correctness, with Acc falling to 10.0. A sharp decline in F-Acc for the dealx and flopx functions is clearly evident in Table~\ref{tab:function_results}.
This indicates that Stage-2 is essential to the core interaction and programming capabilities of the model, especially programming capabilities.
The Mixed-stage involves mixing all the complete interactions and interaction snippets and fine-tuning on them in a single stage. It mixes Stage-2 and Stage-3. 
A significant decrease can be observed in interaction quality in the Mixed-stage setting. Additionally, there is a notable drop in code correctness, with Acc falling to 20.0. 
This indicates that a mixed-stage training strategy for complete interactions and interaction snippets hinders both the interaction and programmig capabilities of the model. This suggests that our three-stage training strategy effectively enhances the joint capability of interaction and programmig of the LLM as a \model{}.

\begin{table}[ht]
    \centering
    \resizebox{\linewidth}{!}{
    \begin{tabular}{p{1.5cm}p{9.6cm}}
    \toprule
    \textbf{Case 1} & \textit{blind: Randomly choose two players as small blind and big blind respectively. Place 1/2 minimum bet for small blind and double minimum bet for big blind.} \\
    \midrule
    \textbf{GPT-4o} & 
    \parbox[t]{9.6cm}{
    \textit{
        def blind(self): \\
    \myindent def bet(player\_id, amount): \\
        \myindent \myindent self.players[player\_id].bet += amount \\
        \myindent \myindent self.players[player\_id].remain -= amount \\
    \myindent small\_blind, big\_blind = \textcolor{red!80!black}{random.choice(self.player\_ids, 2)} \\
    \myindent bet(small\_blind, self.min\_bet // 2) \\
    \myindent bet(big\_blind, self.min\_bet * 2)
    }} \\
    \midrule
    \textbf{ChatGE} & 
    \parbox[t]{9.6cm}{
    \textit{
        def blind(self): \\
    \myindent def bet(player\_id, amount): \\
        \myindent \myindent self.players[player\_id].bet += amount \\
        \myindent \myindent self.players[player\_id].remain -= amount \\
    \myindent small\_blind, big\_blind = \textcolor{green!50!black}{random\_choice(self.player\_ids, 2)} \\
    \myindent bet(small\_blind, self.min\_bet // 2) \\
    \myindent bet(big\_blind, self.min\_bet * 2) 
    }} \\
    \midrule
    \midrule
    \textbf{Case 2} & \textit{dealx: Deal x cards to each player and discard 1 cards from the deck afterward.} \\
    \midrule
    \textbf{GPT-4o} & 
    \parbox[t]{9.6cm}{
    \textit{
        def dealx(self, x): \\
    \myindent for \_ in range(x): \\
        \myindent \myindent for player\_id in self.players: \\
            \myindent \myindent \myindent \textcolor{red!80!black}{self.players[player\_id]['hole'].append(self.deck.pop())} \\
    \myindent self.deck.pop()
    }} \\
    \midrule
    \textbf{ChatGE} & 
    \parbox[t]{9.6cm}{
    \textit{
        def dealx(self, x): \\
    \myindent for i in range(x): \\
        \myindent \myindent for p in self.players: \\
            \myindent \myindent \myindent \textcolor{green!50!black}{self.players[p].hole += [self.deck.pop()]} \\
    \myindent self.deck.pop() 
    }} \\
    \bottomrule
    \end{tabular}
    }
\caption{Case study of the results of GPT-4o and \model{}. Only the code part is retained.}
\label{tab:case-study}
\end{table}

\subsection{Case Study}

In Table~\ref{tab:case-study}, we present two representative cases comparing GPT-4o and \model{}. In Case 1, the code generated by GPT-4o is logically correct, but the function call is used incorrectly. 
The proper usage of ``random.choice'' should be ``random.choice(x)'', but it seems to have confused this with the ``random\_choice'' usage provided in the in-context examples. Similarly, in Case 2, GPT-4o mistakenly treated ``self.players[player\_id]'' as a dict. This can be attributed to its misalignment with the engine, also known as hallucination ~\cite{ji2023survey}. In comparison, \model{} is well-aligned and does not exhibit this phenomenon in the test set.

\section{Conclusion}
This paper introduces the Chat Game Engine (\model{}) and proposes a paradigm for training \model{} to allows users to develop custom games interactively using natural language. To enable an LLM to function as a \model{}, we instruct it to generate script segments, code snippets and interactions for each turn in the development process. 
To facilitate the training process, a data synthesis pipeline is proposed to generate sufficient training data, as well as a three-stage progressive training strategy to enhance the joint capability of interaction and
programming of the LLM. Embodied in a poker game, we demonstrate the performance of the \model{} through a comprehensive evaluation.

\section*{Limitations}
While our \model{} offers exciting potential for applying LLMs as multi-turn game development, several limitations warrant
further exploration: 
(1) Limited scalability: We notice that it is still very hard to generalize \model{} to all games or all game engines. Instead, in this work, we choose a specific game to illustrate the idea of \model{}. The entire data generation and training process must be repeated to adapt this approach to a new game. However, all the prompts we use are designed to be game-agnostic, making them easily adaptable for use in other games. Of course, our future work will definitely focuses on the scalability of \model{}.
(2) Limited scope and modalities: Our current \model{} primarily support text-based games like Poker. Additional modalities such as images, sound, or video could enrich the game and are almost essential in modern video games, but this expansion presents technical and design challenges.
These limitations highlight the importance of ongoing research and development efforts aimed at addressing the challenges associated with LLM-based game development.


\bibliography{custom}

\begin{thebibliography}{48}
\providecommand{\natexlab}[1]{#1}

\bibitem[{Achiam et~al.(2023)Achiam, Adler, Agarwal, Ahmad, Akkaya, Aleman, Almeida, Altenschmidt, Altman, Anadkat et~al.}]{achiam2023gpt}
Josh Achiam, Steven Adler, Sandhini Agarwal, Lama Ahmad, Ilge Akkaya, Florencia~Leoni Aleman, Diogo Almeida, Janko Altenschmidt, Sam Altman, Shyamal Anadkat, et~al. 2023.
\newblock Gpt-4 technical report.
\newblock \emph{arXiv preprint arXiv:2303.08774}.

\bibitem[{Bengio et~al.(2009)Bengio, Louradour, Collobert, and Weston}]{bengio2009curriculum}
Yoshua Bengio, J{\'e}r{\^o}me Louradour, Ronan Collobert, and Jason Weston. 2009.
\newblock Curriculum learning.
\newblock In \emph{Proceedings of the 26th annual international conference on machine learning}, pages 41--48.

\bibitem[{Borisov et~al.(2022)Borisov, Se{\ss}ler, Leemann, Pawelczyk, and Kasneci}]{borisov2022language}
Vadim Borisov, Kathrin Se{\ss}ler, Tobias Leemann, Martin Pawelczyk, and Gjergji Kasneci. 2022.
\newblock Language models are realistic tabular data generators.
\newblock \emph{arXiv preprint arXiv:2210.06280}.

\bibitem[{Brown et~al.(2020)Brown, Mann, Ryder, Subbiah, Kaplan, Dhariwal, Neelakantan, Shyam, Sastry, Askell et~al.}]{brown2020language}
Tom Brown, Benjamin Mann, Nick Ryder, Melanie Subbiah, Jared~D Kaplan, Prafulla Dhariwal, Arvind Neelakantan, Pranav Shyam, Girish Sastry, Amanda Askell, et~al. 2020.
\newblock Language models are few-shot learners.
\newblock \emph{Advances in neural information processing systems}, 33:1877--1901.

\bibitem[{Chintagunta et~al.(2021)Chintagunta, Katariya, Amatriain, and Kannan}]{chintagunta2021medically}
Bharath Chintagunta, Namit Katariya, Xavier Amatriain, and Anitha Kannan. 2021.
\newblock Medically aware gpt-3 as a data generator for medical dialogue summarization.
\newblock In \emph{Machine Learning for Healthcare Conference}, pages 354--372. PMLR.

\bibitem[{Davidson et~al.(2023)Davidson, Romeo, Shu, Gung, Gupta, Mansour, and Zhang}]{davidson2023user}
Sam Davidson, Salvatore Romeo, Raphael Shu, James Gung, Arshit Gupta, Saab Mansour, and Yi~Zhang. 2023.
\newblock User simulation with large language models for evaluating task-oriented dialogue.
\newblock \emph{arXiv preprint arXiv:2309.13233}.

\bibitem[{Dubey et~al.(2024)Dubey, Jauhri, Pandey, Kadian, Al-Dahle, Letman, Mathur, Schelten, Yang, Fan et~al.}]{dubey2024llama}
Abhimanyu Dubey, Abhinav Jauhri, Abhinav Pandey, Abhishek Kadian, Ahmad Al-Dahle, Aiesha Letman, Akhil Mathur, Alan Schelten, Amy Yang, Angela Fan, et~al. 2024.
\newblock The llama 3 herd of models.
\newblock \emph{arXiv preprint arXiv:2407.21783}.

\bibitem[{Eladhari(2018)}]{DBLP:conf/icids/Eladhari18}
Mirjam~Palosaari Eladhari. 2018.
\newblock \href {https://doi.org/10.1007/978-3-030-04028-4\_5} {Re-tellings: The fourth layer of narrative as an instrument for critique}.
\newblock In \emph{Interactive Storytelling - 11th International Conference on Interactive Digital Storytelling, {ICIDS} 2018, Dublin, Ireland, December 5-8, 2018, Proceedings}, volume 11318 of \emph{Lecture Notes in Computer Science}, pages 65--78. Springer.

\bibitem[{Fan et~al.(2022)Fan, Wang, Jiang, Mandlekar, Yang, Zhu, Tang, Huang, Zhu, and Anandkumar}]{DBLP:conf/nips/FanWJMYZTHZA22}
Linxi Fan, Guanzhi Wang, Yunfan Jiang, Ajay Mandlekar, Yuncong Yang, Haoyi Zhu, Andrew Tang, De{-}An Huang, Yuke Zhu, and Anima Anandkumar. 2022.
\newblock \href {http://papers.nips.cc/paper\_files/paper/2022/hash/74a67268c5cc5910f64938cac4526a90-Abstract-Datasets\_and\_Benchmarks.html} {Minedojo: Building open-ended embodied agents with internet-scale knowledge}.
\newblock In \emph{Advances in Neural Information Processing Systems 35: Annual Conference on Neural Information Processing Systems 2022, NeurIPS 2022, New Orleans, LA, USA, November 28 - December 9, 2022}.

\bibitem[{Gallotta et~al.(2024)Gallotta, Todd, Zammit, Earle, Liapis, Togelius, and Yannakakis}]{gallotta2024large}
Roberto Gallotta, Graham Todd, Marvin Zammit, Sam Earle, Antonios Liapis, Julian Togelius, and Georgios~N Yannakakis. 2024.
\newblock Large language models and games: A survey and roadmap.
\newblock \emph{arXiv preprint arXiv:2402.18659}.

\bibitem[{Guo et~al.(2023)Guo, Jin, Liu, Huang, Shi, Yu, Liu, Li, Xiong, Xiong et~al.}]{guo2023evaluating}
Zishan Guo, Renren Jin, Chuang Liu, Yufei Huang, Dan Shi, Linhao Yu, Yan Liu, Jiaxuan Li, Bojian Xiong, Deyi Xiong, et~al. 2023.
\newblock Evaluating large language models: A comprehensive survey.
\newblock \emph{arXiv preprint arXiv:2310.19736}.

\bibitem[{Hu et~al.(2021)Hu, Shen, Wallis, Allen-Zhu, Li, Wang, Wang, and Chen}]{hu2021lora}
Edward~J Hu, Yelong Shen, Phillip Wallis, Zeyuan Allen-Zhu, Yuanzhi Li, Shean Wang, Lu~Wang, and Weizhu Chen. 2021.
\newblock Lora: Low-rank adaptation of large language models.
\newblock \emph{arXiv preprint arXiv:2106.09685}.

\bibitem[{Ji et~al.(2023)Ji, Lee, Frieske, Yu, Su, Xu, Ishii, Bang, Madotto, and Fung}]{ji2023survey}
Ziwei Ji, Nayeon Lee, Rita Frieske, Tiezheng Yu, Dan Su, Yan Xu, Etsuko Ishii, Ye~Jin Bang, Andrea Madotto, and Pascale Fung. 2023.
\newblock Survey of hallucination in natural language generation.
\newblock \emph{ACM Computing Surveys}, 55(12):1--38.

\bibitem[{Jin et~al.(2023)Jin, Han, Yang, Jiang, Chang, and Hu}]{jin2023growlength}
Hongye Jin, Xiaotian Han, Jingfeng Yang, Zhimeng Jiang, Chia-Yuan Chang, and Xia Hu. 2023.
\newblock Growlength: Accelerating llms pretraining by progressively growing training length.
\newblock \emph{arXiv preprint arXiv:2310.00576}.

\bibitem[{K{\"{u}}ttler et~al.(2020)K{\"{u}}ttler, Nardelli, Miller, Raileanu, Selvatici, Grefenstette, and Rockt{\"{a}}schel}]{DBLP:conf/nips/KuttlerNMRSGR20}
Heinrich K{\"{u}}ttler, Nantas Nardelli, Alexander~H. Miller, Roberta Raileanu, Marco Selvatici, Edward Grefenstette, and Tim Rockt{\"{a}}schel. 2020.
\newblock \href {https://proceedings.neurips.cc/paper/2020/hash/569ff987c643b4bedf504efda8f786c2-Abstract.html} {The nethack learning environment}.
\newblock In \emph{Advances in Neural Information Processing Systems 33: Annual Conference on Neural Information Processing Systems 2020, NeurIPS 2020, December 6-12, 2020, virtual}.

\bibitem[{Liu et~al.(2024)Liu, Yan, Zaharia, and Abbeel}]{liu2024world}
Hao Liu, Wilson Yan, Matei Zaharia, and Pieter Abbeel. 2024.
\newblock World model on million-length video and language with blockwise ringattention.
\newblock \emph{arXiv preprint arXiv:2402.08268}.

\bibitem[{Lowe et~al.(2020)Lowe, Gupta, Foerster, Kiela, and Pineau}]{DBLP:conf/iclr/Lowe0FKP20}
Ryan Lowe, Abhinav Gupta, Jakob~N. Foerster, Douwe Kiela, and Joelle Pineau. 2020.
\newblock \href {https://openreview.net/forum?id=rJxGLlBtwH} {On the interaction between supervision and self-play in emergent communication}.
\newblock In \emph{8th International Conference on Learning Representations, {ICLR} 2020, Addis Ababa, Ethiopia, April 26-30, 2020}. OpenReview.net.

\bibitem[{Luo et~al.(2024)Luo, Tang, Wang, and Zhang}]{luo2024duetsim}
Xiang Luo, Zhiwen Tang, Jin Wang, and Xuejie Zhang. 2024.
\newblock Duetsim: Building user simulator with dual large language models for task-oriented dialogues.
\newblock In \emph{Proceedings of the 2024 Joint International Conference on Computational Linguistics, Language Resources and Evaluation (LREC-COLING 2024)}, pages 5414--5424.

\bibitem[{Mnih et~al.(2013)Mnih, Kavukcuoglu, Silver, Graves, Antonoglou, Wierstra, and Riedmiller}]{DBLP:journals/corr/MnihKSGAWR13}
Volodymyr Mnih, Koray Kavukcuoglu, David Silver, Alex Graves, Ioannis Antonoglou, Daan Wierstra, and Martin~A. Riedmiller. 2013.
\newblock \href {https://arxiv.org/abs/1312.5602} {Playing atari with deep reinforcement learning}.
\newblock \emph{CoRR}, abs/1312.5602.

\bibitem[{Ouyang et~al.(2022)Ouyang, Wu, Jiang, Almeida, Wainwright, Mishkin, Zhang, Agarwal, Slama, Ray et~al.}]{ouyang2022training}
Long Ouyang, Jeffrey Wu, Xu~Jiang, Diogo Almeida, Carroll Wainwright, Pamela Mishkin, Chong Zhang, Sandhini Agarwal, Katarina Slama, Alex Ray, et~al. 2022.
\newblock Training language models to follow instructions with human feedback.
\newblock \emph{Advances in neural information processing systems}, 35:27730--27744.

\bibitem[{Peng et~al.(2023)Peng, Li, He, Galley, and Gao}]{peng2023instruction}
Baolin Peng, Chunyuan Li, Pengcheng He, Michel Galley, and Jianfeng Gao. 2023.
\newblock Instruction tuning with gpt-4.
\newblock \emph{arXiv preprint arXiv:2304.03277}.

\bibitem[{Raffel et~al.(2020)Raffel, Shazeer, Roberts, Lee, Narang, Matena, Zhou, Li, and Liu}]{raffel2020exploring}
Colin Raffel, Noam Shazeer, Adam Roberts, Katherine Lee, Sharan Narang, Michael Matena, Yanqi Zhou, Wei Li, and Peter~J Liu. 2020.
\newblock Exploring the limits of transfer learning with a unified text-to-text transformer.
\newblock \emph{Journal of machine learning research}, 21(140):1--67.

\bibitem[{Ranella and Eger(2023)}]{DBLP:conf/exag/RanellaE23}
Noah Ranella and Markus Eger. 2023.
\newblock \href {https://ceur-ws.org/Vol-3626/paper7.pdf} {Towards automated video game commentary using generative {AI}}.
\newblock In \emph{Proceedings of the Experimental Artificial Intelligence in Games Workshop co-located with the 19th {AAAI} Conference on Artificial Intelligence and Interactive Digital Entertainment {(AIIDE} 2023), Salt Lake City, Utah, USA, October 8, 2023}, volume 3626 of \emph{{CEUR} Workshop Proceedings}. CEUR-WS.org.

\bibitem[{Schick and Sch{\"u}tze(2021)}]{schick2021generating}
Timo Schick and Hinrich Sch{\"u}tze. 2021.
\newblock Generating datasets with pretrained language models.
\newblock In \emph{Proceedings of the 2021 Conference on Empirical Methods in Natural Language Processing}, pages 6943--6951.

\bibitem[{Sekuli{\'c} et~al.(2024)Sekuli{\'c}, Terragni, Guimar{\~a}es, Khau, Guedes, Filipavicius, Manso, and Mathis}]{sekulic2024reliable}
Ivan Sekuli{\'c}, Silvia Terragni, Victor Guimar{\~a}es, Nghia Khau, Bruna Guedes, Modestas Filipavicius, Andre~Ferreira Manso, and Roland Mathis. 2024.
\newblock Reliable llm-based user simulator for task-oriented dialogue systems.
\newblock In \emph{Proceedings of the 1st Workshop on Simulating Conversational Intelligence in Chat (SCI-CHAT 2024)}, pages 19--35.

\bibitem[{Shanahan et~al.(2023)Shanahan, McDonell, and Reynolds}]{DBLP:journals/nature/ShanahanMR23}
Murray Shanahan, Kyle McDonell, and Laria Reynolds. 2023.
\newblock \href {https://doi.org/10.1038/S41586-023-06647-8} {Role play with large language models}.
\newblock \emph{Nat.}, 623(7987):493--498.

\bibitem[{Shao et~al.(2023{\natexlab{a}})Shao, Li, Dai, and Qiu}]{shao2023character}
Yunfan Shao, Linyang Li, Junqi Dai, and Xipeng Qiu. 2023{\natexlab{a}}.
\newblock Character-llm: A trainable agent for role-playing.
\newblock \emph{arXiv preprint arXiv:2310.10158}.

\bibitem[{Shao et~al.(2023{\natexlab{b}})Shao, Gong, Shen, Huang, Duan, and Chen}]{shao2023synthetic}
Zhihong Shao, Yeyun Gong, Yelong Shen, Minlie Huang, Nan Duan, and Weizhu Chen. 2023{\natexlab{b}}.
\newblock Synthetic prompting: Generating chain-of-thought demonstrations for large language models.
\newblock In \emph{International Conference on Machine Learning}, pages 30706--30775. PMLR.

\bibitem[{Sun et~al.(2024)Sun, Shen, Zhou, Zhang, Chen, Cox, Yang, and Gan}]{sun2024principle}
Zhiqing Sun, Yikang Shen, Qinhong Zhou, Hongxin Zhang, Zhenfang Chen, David Cox, Yiming Yang, and Chuang Gan. 2024.
\newblock Principle-driven self-alignment of language models from scratch with minimal human supervision.
\newblock \emph{Advances in Neural Information Processing Systems}, 36.

\bibitem[{Terragni et~al.(2023)Terragni, Filipavicius, Khau, Guedes, Manso, and Mathis}]{terragni2023context}
Silvia Terragni, Modestas Filipavicius, Nghia Khau, Bruna Guedes, Andr{\'e} Manso, and Roland Mathis. 2023.
\newblock In-context learning user simulators for task-oriented dialog systems.
\newblock \emph{arXiv preprint arXiv:2306.00774}.

\bibitem[{Touvron et~al.(2023)Touvron, Martin, Stone, Albert, Almahairi, Babaei, Bashlykov, Batra, Bhargava, Bhosale et~al.}]{touvron2023llama}
Hugo Touvron, Louis Martin, Kevin Stone, Peter Albert, Amjad Almahairi, Yasmine Babaei, Nikolay Bashlykov, Soumya Batra, Prajjwal Bhargava, Shruti Bhosale, et~al. 2023.
\newblock Llama 2: Open foundation and fine-tuned chat models.
\newblock \emph{arXiv preprint arXiv:2307.09288}.

\bibitem[{Uludagli and Oguz(2023)}]{DBLP:journals/air/UludagliO23}
Muhtar~{\c{C}}agkan Uludagli and Kaya Oguz. 2023.
\newblock \href {https://doi.org/10.1007/S10462-023-10491-7} {Non-player character decision-making in computer games}.
\newblock \emph{Artif. Intell. Rev.}, 56(12):14159--14191.

\bibitem[{Vakil and Amiri(2023)}]{vakil2023complexity}
Nidhi Vakil and Hadi Amiri. 2023.
\newblock Complexity-guided curriculum learning for text graphs.
\newblock \emph{arXiv preprint arXiv:2311.13472}.

\bibitem[{Valencia-Garc{\'\i}a et~al.(2016)Valencia-Garc{\'\i}a, Lagos-Ortiz, Alcaraz-M{\'a}rmol, del Cioppo, and Vera-Lucio}]{valencia2016technologies}
Rafael Valencia-Garc{\'\i}a, Katty Lagos-Ortiz, Gema Alcaraz-M{\'a}rmol, Javier del Cioppo, and Nestor Vera-Lucio. 2016.
\newblock Technologies and innovation: Second international conference, citi 2016, guayaquil, ecuador, november 23-25, 2016.
\newblock \emph{Proceedings. Communications in Computer and Information Science}, 658.

\bibitem[{Vinyals et~al.(2019)Vinyals, Babuschkin, Czarnecki, Mathieu, Dudzik, Chung, Choi, Powell, Ewalds, Georgiev, Oh, Horgan, Kroiss, Danihelka, Huang, Sifre, Cai, Agapiou, Jaderberg, Vezhnevets, Leblond, Pohlen, Dalibard, Budden, Sulsky, Molloy, Paine, G{\"{u}}l{\c{c}}ehre, Wang, Pfaff, Wu, Ring, Yogatama, W{\"{u}}nsch, McKinney, Smith, Schaul, Lillicrap, Kavukcuoglu, Hassabis, Apps, and Silver}]{DBLP:journals/nature/VinyalsBCMDCCPE19}
Oriol Vinyals, Igor Babuschkin, Wojciech~M. Czarnecki, Micha{\"{e}}l Mathieu, Andrew Dudzik, Junyoung Chung, David~H. Choi, Richard Powell, Timo Ewalds, Petko Georgiev, Junhyuk Oh, Dan Horgan, Manuel Kroiss, Ivo Danihelka, Aja Huang, Laurent Sifre, Trevor Cai, John~P. Agapiou, Max Jaderberg, Alexander~Sasha Vezhnevets, R{\'{e}}mi Leblond, Tobias Pohlen, Valentin Dalibard, David Budden, Yury Sulsky, James Molloy, Tom~Le Paine, {\c{C}}aglar G{\"{u}}l{\c{c}}ehre, Ziyu Wang, Tobias Pfaff, Yuhuai Wu, Roman Ring, Dani Yogatama, Dario W{\"{u}}nsch, Katrina McKinney, Oliver Smith, Tom Schaul, Timothy~P. Lillicrap, Koray Kavukcuoglu, Demis Hassabis, Chris Apps, and David Silver. 2019.
\newblock \href {https://doi.org/10.1038/S41586-019-1724-Z} {Grandmaster level in starcraft {II} using multi-agent reinforcement learning}.
\newblock \emph{Nat.}, 575(7782):350--354.

\bibitem[{Wang et~al.(2023{\natexlab{a}})Wang, Xie, Jiang, Mandlekar, Xiao, Zhu, Fan, and Anandkumar}]{DBLP:journals/corr/abs-2305-16291}
Guanzhi Wang, Yuqi Xie, Yunfan Jiang, Ajay Mandlekar, Chaowei Xiao, Yuke Zhu, Linxi Fan, and Anima Anandkumar. 2023{\natexlab{a}}.
\newblock \href {https://doi.org/10.48550/ARXIV.2305.16291} {Voyager: An open-ended embodied agent with large language models}.
\newblock \emph{CoRR}, abs/2305.16291.

\bibitem[{Wang et~al.(2023{\natexlab{b}})Wang, Wang, Liu, Chen, Yuan, Peng, and Ji}]{wang2023mint}
Xingyao Wang, Zihan Wang, Jiateng Liu, Yangyi Chen, Lifan Yuan, Hao Peng, and Heng Ji. 2023{\natexlab{b}}.
\newblock Mint: Evaluating llms in multi-turn interaction with tools and language feedback.
\newblock \emph{arXiv preprint arXiv:2309.10691}.

\bibitem[{Wang et~al.(2023{\natexlab{c}})Wang, Yu, Zeng, Yang, Wang, Chen, Jiang, Xie, Wang, Xie et~al.}]{wang2023pandalm}
Yidong Wang, Zhuohao Yu, Zhengran Zeng, Linyi Yang, Cunxiang Wang, Hao Chen, Chaoya Jiang, Rui Xie, Jindong Wang, Xing Xie, et~al. 2023{\natexlab{c}}.
\newblock Pandalm: An automatic evaluation benchmark for llm instruction tuning optimization.
\newblock \emph{arXiv preprint arXiv:2306.05087}.

\bibitem[{Wang et~al.(2022)Wang, Kordi, Mishra, Liu, Smith, Khashabi, and Hajishirzi}]{wang2022self}
Yizhong Wang, Yeganeh Kordi, Swaroop Mishra, Alisa Liu, Noah~A Smith, Daniel Khashabi, and Hannaneh Hajishirzi. 2022.
\newblock Self-instruct: Aligning language models with self-generated instructions.
\newblock \emph{arXiv preprint arXiv:2212.10560}.

\bibitem[{Wei et~al.(2022)Wei, Wang, Schuurmans, Bosma, Xia, Chi, Le, Zhou et~al.}]{wei2022chain}
Jason Wei, Xuezhi Wang, Dale Schuurmans, Maarten Bosma, Fei Xia, Ed~Chi, Quoc~V Le, Denny Zhou, et~al. 2022.
\newblock Chain-of-thought prompting elicits reasoning in large language models.
\newblock \emph{Advances in neural information processing systems}, 35:24824--24837.

\bibitem[{Wu et~al.(2023)Wu, Liu, Zhao, and Zhang}]{wu2023empower}
Hongqiu Wu, Linfeng Liu, Hai Zhao, and Min Zhang. 2023.
\newblock Empower nested boolean logic via self-supervised curriculum learning.
\newblock \emph{arXiv preprint arXiv:2310.05450}.

\bibitem[{Wu et~al.(2024{\natexlab{a}})Wu, Liu, Wang, and Zhao}]{wu2024instruction}
Hongqiu Wu, Xingyuan Liu, Yan Wang, and Hai Zhao. 2024{\natexlab{a}}.
\newblock Instruction-driven game engine: A poker case study.
\newblock In \emph{Proceedings of the 2024 Conference on Empirical Methods in Natural Language Processing: System Demonstrations}, pages 507--519.

\bibitem[{Wu et~al.(2024{\natexlab{b}})Wu, Wu, Jiang, Liu, Hong, Zhao, and Zhang}]{wu2024role}
Weiqi Wu, Hongqiu Wu, Lai Jiang, Xingyuan Liu, Jiale Hong, Hai Zhao, and Min Zhang. 2024{\natexlab{b}}.
\newblock From role-play to drama-interaction: An llm solution.
\newblock \emph{arXiv preprint arXiv:2405.14231}.

\bibitem[{Xiang et~al.(2024)Xiang, Zhu, Lou, Chen, Pan, Jin, Chen, and Sun}]{xiang2024simuser}
Wei Xiang, Hanfei Zhu, Suqi Lou, Xinli Chen, Zhenghua Pan, Yuping Jin, Shi Chen, and Lingyun Sun. 2024.
\newblock Simuser: Generating usability feedback by simulating various users interacting with mobile applications.
\newblock In \emph{Proceedings of the CHI Conference on Human Factors in Computing Systems}, pages 1--17.

\bibitem[{Xu et~al.(2023)Xu, Wang, Li, Luo, Wang, Liu, and Liu}]{DBLP:journals/corr/abs-2309-04658}
Yuzhuang Xu, Shuo Wang, Peng Li, Fuwen Luo, Xiaolong Wang, Weidong Liu, and Yang Liu. 2023.
\newblock \href {https://doi.org/10.48550/ARXIV.2309.04658} {Exploring large language models for communication games: An empirical study on werewolf}.
\newblock \emph{CoRR}, abs/2309.04658.

\bibitem[{Yu et~al.(2024{\natexlab{a}})Yu, Zhuang, Zhang, Meng, Ratner, Krishna, Shen, and Zhang}]{yu2024large}
Yue Yu, Yuchen Zhuang, Jieyu Zhang, Yu~Meng, Alexander~J Ratner, Ranjay Krishna, Jiaming Shen, and Chao Zhang. 2024{\natexlab{a}}.
\newblock Large language model as attributed training data generator: A tale of diversity and bias.
\newblock \emph{Advances in Neural Information Processing Systems}, 36.

\bibitem[{Yu et~al.(2024{\natexlab{b}})Yu, Gao, Yao, Wang, Ye, Wang, Xie, Zhang, and Zhang}]{yu2024kieval}
Zhuohao Yu, Chang Gao, Wenjin Yao, Yidong Wang, Wei Ye, Jindong Wang, Xing Xie, Yue Zhang, and Shikun Zhang. 2024{\natexlab{b}}.
\newblock \href {https://doi.org/10.18653/v1/2024.acl-long.325} {{KIE}val: A knowledge-grounded interactive evaluation framework for large language models}.
\newblock In \emph{Proceedings of the 62nd Annual Meeting of the Association for Computational Linguistics (Volume 1: Long Papers)}, pages 5967--5985, Bangkok, Thailand. Association for Computational Linguistics.

\bibitem[{Zheng et~al.(2024)Zheng, Chiang, Sheng, Zhuang, Wu, Zhuang, Lin, Li, Li, Xing et~al.}]{zheng2024judging}
Lianmin Zheng, Wei-Lin Chiang, Ying Sheng, Siyuan Zhuang, Zhanghao Wu, Yonghao Zhuang, Zi~Lin, Zhuohan Li, Dacheng Li, Eric Xing, et~al. 2024.
\newblock Judging llm-as-a-judge with mt-bench and chatbot arena.
\newblock \emph{Advances in Neural Information Processing Systems}, 36.

\end{thebibliography}

\appendix
\section{Ablation statistics} \label{app-sec:ablation-statistics}

\begin{table}[ht]
\resizebox{\linewidth}{!}{
\centering
\begin{tabular}{l|cc|cc}
  & \multicolumn{2}{c|}{\textbf{seed}} & \multicolumn{2}{c}{\textbf{synthesis}}   \\
  & \textit{Com.} & \textit{Snip.} & \textit{Com.}  & 
  \textit{Snip.} \\ 
  & \textit{(20)} & \textit{(180))} & \textit{(36)}  & \textit{(3718)} \\ 
\midrule
\model{} & \checkmark & \checkmark & \checkmark & \checkmark \\
\quad \textit{w/o. $P_{script}$} & \checkmark & \checkmark & \checkmark & \checkmark \\
\quad \textit{w/o. synthesis} & \checkmark & \checkmark &  &  \\
\quad \textit{w/o. Stage-2} & \checkmark & & \checkmark & \\
\quad \textit{w. Mixed-stage} & \checkmark & \checkmark & \checkmark & \checkmark \\
\end{tabular}
}
\caption{Traning data statistics of ablations. \textit{Com.} refers to complete interactions and \textit{Snip.} refers to interaction snippets.}
\label{app-tab:ablation-statistics}
\end{table}

\section{Prompts Demonstration}

\lstset{
  basicstyle=\ttfamily\small,
  breaklines=true,
  captionpos=b,
  columns=fullflexible,
  keepspaces=true,
}

In this section, we provide the prompts used in the paper.
Each \{...\} component above will be substituted with corresponding information. For more details, please refer to our code. These prompts are not designed for any specific game, they can be used to build a \model{} for any \{Game\_name\}.

Table \ref{app-tab:pairs}-\ref{app-tab:complete} present the prompts used in the Data Generation.
Table \ref{app-tab:baselines}-\ref{app-tab:ige} present the system prompts for models.
Table \ref{app-tab:interactor}-\ref{app-tab:evaluator} present the prompts used in the Evaluation.

\begin{table*}[ht]
{\small
    \begin{tcolorbox}[colback=white!95!gray,colframe=gray!50!black,rounded corners, title={Prompt for New Script-code Pairs Generation}]
Generate new code snippet and the corresponding script segment of a \{Game\_name\} game based on the given original code snippet and the corresponding original script segment. \\

1. Modify the code logic to obtain a new code segment and output the corresponding script segment. \\
2. The new code snippet is obtained by modifying the original code snippet. \\
3. Keep the input parameters unchanged, do not introduce new input parameters. \\
4. The generated new code snippet should not introduce new instance attributes and involved methods such as `self.xxx` or `self.xxx\(...\)` compared to the original code snippet. The generated new code snippet can only include instance attributes and instance methods involved in the original code snippet. You cannot create new ones. For example, there is a original code snippet below:
\begin{verbatim}
def bet_done(self, wait_to_bet):
    all_bet = [self.players[p].bet for p in self.get_unfold_players()]
    if not wait_to_bet and all([b == all_bet[0] for b in all_bet]):
        return True
    return False
\end{verbatim}

In this code snippet, the instance attributes and instance methods involved are only `self.players` and `self.get\_unfold\_players()`. Therefore, in the new code snippet generated from this original code snippet, the instance attributes and instance methods involved should also only be `self.players` and `self.get\_unfold\_players()`, other created ones such as `self.group`, `self.discard\_pile`, `self.burn\_pile`, `self.burn\_card` are not allowed to be used. \\
5. Do not use `print` or logging information. \\
6. The script segment can be seen as a description of the code snippet. \\
7. Try to be creative and diverse. \\
8. The output format should follow the original, without any redundant information. \\

------------------------------ \\ 
\textbf{\# Examples} \\
\{In-context Examples\} \\
------------------------------ \\ 

\textbf{\# Start of Official Requests} \\
\textbf{\#\# original code snippet:} \\
\{original\_code\} \\

\textbf{\#\# original script segment:} \\
\{original\_script\}
    \end{tcolorbox}
\caption{Prompt for generating new pairs in Data Generation for \model{}.}
\label{app-tab:pairs}  
}
\end{table*}

\begin{table*}[ht]
{\small
    \begin{tcolorbox}[colback=white!95!gray,colframe=gray!50!black,rounded corners, title={Prompt for Interaction Snippets Generation}]
Generate a dialogue between a user and an assistant based on the following rules and given script segment and code snippet. \\

1. The user edits game script segments using natural language during interactions with the assistant. \\
2. The assistant interacts with the user to achieve interactive game development. The assistant guides the user in editing game script segments, generates corresponding code snippets, and interacts with the user through dialogue. \\
3. Each turn of the assistant's output should include three processes: "script", "code", and "utter", corresponding to three blocks:  <script></script>, <code></code>, <utter></utter>. Formally, these three blocks must exist, even if the content is empty. \\
4. The 'script' process: The assistant generates the game script segment based on the user's input of the current turn. Return modifications to the script as changes, rather than returning the entire script. The script is a Python dict, so you can use simple Python code to represent modifications to it, such as: script['xxx'] = 'xxx'. The 'script' process should be enclosed using '<script>' tag. \\
5. The 'code' process: The assistant generates the corresponding Python code snippet based on the game script segment from the 'script' process. The complete code is a CustomGame class that inherits from GameBase class, but only the methods related to the given script segment need to be generated. The 'code' process should be enclosed using '<code>' tag. \\
6. The 'utter' process: The assistant interacts with the user, including responding to the user's input of the current turn, summarizing the results of the current turn, and guiding the user to continue with the next turn of interaction. The 'utter' process should be enclosed using '<utter>' tag. \\
7. The script segment and code snippet have already been provided. In the assistant's 'script' and 'code' process, use the given script segment and code snippet; do not write your own. \\
8. The assistant does not know about the existence of the script segment in the dialogue and needs to obtain it from the user's input. \\
9. The given script segment and code snippet are essentially an outline of the plot development. The assistant's 'script' and 'code' process must be entirely derived from or inferred from the user's input. The user's input should be more natural language-based and not a direct copy of the given script segement. \\
10. The dialogue must cover and only cover the given script segment, and no other content should appear. \\

\{Formatting Instruction\} \\

------------------------------ \\ 
\textbf{\# Examples} \\
\{In-context Examples\} \\
------------------------------ \\ 

\textbf{\# Start of Official Requests} \\
\textbf{\#\# script segment:} \\
\{script segment\} \\

\textbf{\#\# code snippet:} \\
\{code snippet\} \\

\textbf{\#\# dialogue:}
    \end{tcolorbox}
\caption{Prompt for generating interaction snippets in Data Generation for \model{}.}
\label{app-tab:snippets}  
}
\end{table*}

\begin{table*}[ht]
{\small
    \begin{tcolorbox}[colback=white!95!gray,colframe=gray!50!black,rounded corners, title={Prompt for Complete Interactions Generation}]
Generate a dialogue between a user and an assistant based on the following rules and given script segment and code snippet. \\

1. The user edits game script segments using natural language during interactions with the assistant. \\
2. The assistant interacts with the user to achieve interactive game development. The assistant guides the user in editing game script segments, generates corresponding code snippets, and interacts with the user through dialogue. \\
3. Each turn of the assistant's output should include three processes: "script", "code", and "utter", corresponding to three blocks:  <script></script>, <code></code>, <utter></utter>. Formally, these three blocks must exist, even if the content is empty. \\
4. The 'script' process: The assistant generates the game script segment based on the user's input of the last turn. Return modifications to the script as changes, rather than returning the entire script. The script is a Python dict, so you can use simple Python code to represent modifications to it, such as: script['xxx'] = 'xxx'. The 'script' process should be enclosed using '<script>' tag. \\
5. The 'code' process: The assistant generates the corresponding Python code snippet based on the game script segment from the 'script' process. The complete code is a CustomGame class that inherits from GameBase class, but only the methods related to the given script segment need to be generated. The 'code' process should be enclosed using '<code>' tag. \\
6. The 'utter' process: The assistant interacts with the user, including responding to the user's input of the last turn, summarizing the results of the current turn, and guiding the user to continue with the current turn of interaction. The 'utter' process should be enclosed using '<utter>' tag. \\
7. The script segment and code snippet have already been provided. You need to randomly distribute them across multiple turns and generate an interactive dialogue between the assistant and the user. This means the assistant guides the user step by step to complete this game script segment. In a single turn of dialogue, the user's input should not contain too much information. If a large input is required, it should be divided into multiple turns. \\
8. In the assistant's 'script' and 'code' process, use the given script segment and code snippet; do not write your own. \\
9. The dialogue must cover and only cover all the given script segment, and no other content should appear. \\
10. The assistant does not know about the existence of the script segment in the dialogue and needs to obtain it from the user's input. \\
11. The given script segment and code snippet are essentially an outline of the plot development. The assistant's 'script' and 'code' process must be entirely derived from or inferred from the user's input. The user's input should be more natural language-based and not a direct copy of the given script segement. \\
12. In the first turn, the 'script' and 'code' process of the assistant should be empty because the user has not yet input a game script segment. In the first turn, the assistant should greet the user and start guiding them. In the end, after the user has completed the entire script under the assistant's guidance, the assistant should convey to the user that the game development is complete. \\
13. The assistant should guide the user step by step along a specific line to complete each part of the game script: \\
\{Game\_script\_line\} \\

\{Formatting Instruction\} \\

------------------------------ \\ 
\textbf{\# Examples} \\
\{In-context Examples\} \\
------------------------------ \\ 

\textbf{\# Start of Official Requests} \\
\textbf{\#\# script segment:} \\
\{script segment\} \\

\textbf{\#\# code snippet:} \\
\{code snippet\} \\

\textbf{\#\# dialogue:}
    \end{tcolorbox}
\caption{Prompt for generating complete interactions in Data Generation for \model{}.}
\label{app-tab:complete}
}
\end{table*}

\begin{table*}[ht]
{\small
    \begin{tcolorbox}[colback=white!95!gray,colframe=gray!50!black,rounded corners, title={System Prompt for Baselines in a 5-shot Setting}]
You are a helpful assistant assigned to interact with the user for the interactive development of a \{Game\_name\} game. \\

1. The user edits game script segments using natural language. \\
2. The assistant guides the user in editing game script segments, generates corresponding code snippets, and interacts with the user through dialogue. \\
3. Each turn of the assistant's output should include three processes: "script", "code", and "utter", corresponding to three blocks:  <script></script>, <code></code>, <utter></utter>. Formally, these three blocks must exist, even if the content is empty. \\
4. The 'script' process: The assistant generates the game script segment based on the user's input of the current turn. Return modifications to the script as changes, rather than returning the entire script. The script is a existing Python dict, so you can use simple Python code to represent modifications to it, such as: script['xxx'] = 'xxx'. The 'script' process should be enclosed using '<script>' tag. \\
5. The 'code' process: The assistant generates the corresponding Python code snippet based on the game script segment from the 'script' process. The complete code is a CustomGame class that inherits from GameBase class, but only the methods related to the given script segment need to be generated. The 'code' process should be enclosed using '<code>' tag. \\
6. The 'utter' process: The assistant interacts with the user, including responding to the user's input of the current turn, summarizing the results of the current turn, and guiding the user to continue with the next turn of interaction. The 'utter' process should be enclosed using '<utter>' tag. \\
7. The assistant's 'script' and 'code' process must be entirely derived from or inferred from the user's input. If the user's input lacks the required information, ask the user for further details, and both the 'script' process and the 'code' process of the assistant should be empty. \\
8. If the user's input is unrelated to the script or insufficient to cause changes in the script, the 'script' process and the 'code' process of the assistant should both be empty. \\
9. If the user has any questions, answer them instead of randomly modifying the script and code on your own. \\
10. In the first turn, the 'script' and 'code' process of the assistant should be empty because the user has not yet input a game script segment. In the first turn, the assistant should greet the user and start guiding them. In the end, after the user has completed the entire script under the assistant's guidance, the assistant should convey to the user that the game development is complete. \\
11. The assistant should guide the user step by step along a specific line to complete each part of the game script, referring to the given script template. \\

\{Formatting Instruction\} \\

\textbf{\# script template} \\
\{script template\} \\

------------------------------ \\ 
\textbf{\# Examples} \\
\{In-context Examples\} \\
------------------------------ 
    \end{tcolorbox}
\caption{System prompt for baselines in a 5-shot Setting.}
\label{app-tab:baselines}
}
\end{table*}

\begin{table*}[ht]
{\small
    \begin{tcolorbox}[colback=white!95!gray,colframe=gray!50!black,rounded corners, title={System Prompt for \model{}}]
You are a helpful assistant assigned to interact with the user for the interactive development of a \{Game\_name\} game. \\

1. The user edits game script segments using natural language. \\
2. The assistant guides the user in editing game script segments, generates corresponding code snippets, and interacts with the user through dialogue. \\
3. Each turn of the assistant's output should include three processes: "script", "code", and "utter", corresponding to three blocks:  <script></script>, <code></code>, <utter></utter>. Formally, these three blocks must exist, even if the content is empty. \\
4. The 'script' process: The assistant generates the game script segment based on the user's input of the current turn. Return modifications to the script as changes, rather than returning the entire script. The script is a existing Python dict, so you can use simple Python code to represent modifications to it, such as: script['xxx'] = 'xxx'. The 'script' process should be enclosed using '<script>' tag. \\
5. The 'code' process: The assistant generates the corresponding Python code snippet based on the game script segment from the 'script' process. The complete code is a CustomGame class that inherits from GameBase class, but only the methods related to the given script segment need to be generated. The 'code' process should be enclosed using '<code>' tag. \\
6. The 'utter' process: The assistant interacts with the user, including responding to the user's input of the current turn, summarizing the results of the current turn, and guiding the user to continue with the next turn of interaction. The 'utter' process should be enclosed using '<utter>' tag. \\
7. The assistant's 'script' and 'code' process must be entirely derived from or inferred from the user's input. If the user's input lacks the required information, ask the user for further details, and both the 'script' process and the 'code' process of the assistant should be empty. \\
8. If the user's input is unrelated to the script or insufficient to cause changes in the script, the 'script' process and the 'code' process of the assistant should both be empty. \\
9. If the user has any questions, answer them instead of randomly modifying the script and code on your own.
    \end{tcolorbox}
\caption{System prompt for \model{}.}
\label{app-tab:ige}
}
\end{table*}

\begin{table*}[ht]
{\small
    \begin{tcolorbox}[colback=white!95!gray,colframe=gray!50!black,rounded corners, title={System Prompt for Interactor}]
You are a user (as in the Example) of an interactive \{Game\_name\} game development application of a \{Game\_name\} game, interacting with me (the assistant). \\

1. You should attempt to use natural language to edit game script segments. \\
2. You should focus on the "utter" part enclosed by the <utter></utter> tag in my output and interact with it according to its guidance. \\
3. Your response does not need to include any tags. \\
4. A game script will be given. Assume this is the game script you have in mind. You need to interactively present your ideas under the guidance of the me step by step, i.e., respond based on the relevant parts of the given script. Try not to output too much in one turn. \\
5. Try to use natural language instead of directly copying the given script segments. \\
6. Your responses should be as concise as possible and should not include the thought process. \\

------------------------------ \\ 
\textbf{\# Examples} \\
\{In-context Examples\} \\
------------------------------ \\

\textbf{\# Start of Official Requests} \\
\textbf{\#\# given game script:} \\
\{game script\}
    \end{tcolorbox}
\caption{System prompt for interactor in evaluation.}
\label{app-tab:interactor}
}
\end{table*}

\begin{table*}[ht]
{\small
    \begin{tcolorbox}[colback=white!95!gray,colframe=gray!50!black,rounded corners, title={System Prompt for Evaluator}]
You are an objective evaluator in an interview. Your task is to evaluate a assistant's performance during a series of interactions with an user. The conversation alternates between the user (marked with 'user:') and the assistant (marked with 'assistant'). Evaluate the assistant's performance in the interactions as well as in context, based on the following aspects independently, rating each on a scale from 1 (Poor) to 4 (Good): \\

Guidance: How the response guide the user step-by-step to complete the game. \\
Logic: Logical structure and soundness of reasoning, including the support and validity of conclusions. Whether conclusions are well-supported and arguments are free from logical fallacies. \\
Relevance: How the response relates to the topic. Ensure responses are within the scope of the "assistant" role, avoiding unpermitted role shifts. \\
Coherence: How well the response integrates into the context. Consistency with previous statements and overall conversational flow. \\
Conciseness: Brevity and clarity of the response. Clear, to-the-point communication, free from extraneous elaboration or repetitive words. \\

\textbf{Scoring Guide:} \\
1 (Poor): Significant deficiencies or inaccuracies in the aspect. \\
2 (Below Average): Noticeable weaknesses, partially on target but lacking in several areas. \\
3 (Above Average): Solid and competent, mostly on target with only a few minor shortcomings. \\
4 (Good): Strong performance, fully meets and often surpasses expectations. \\

\textbf{Evaluation Rules:} \\
1. Evaluate the assistant consistently and objectively without bias, strictly adhering to scoring guide. \\
2. Score from 1 to 4 for each aspect independently, using only integers. Low score in one aspect should not influence another aspect. Write a brief comment before scoring in the JSON output structure. \\
3. Write a overall comment and then give an overall score (same scoring guide). The overall comment should be brief and clear. Consider the performance throughout the interaction, not just in the latest round. \\
4. Format of Evaluation: Output in JSON format strictly following the template, without any other words: \\
{guidance": {"comment": "", "score": 0}, "logic": {"comment": "", "score": 0}, "relevance": {"comment": "", "score": 0}, "coherence": {"comment": "", "score": 0}, "conciseness": {"comment": "", "score": 0}, "overall": {"comment": "", "score": 0}} \\

\textbf{\# interactions:} \\
\{dialogue\} \\

\textbf{\# Evaluation:}
    \end{tcolorbox}
\caption{System prompt for evaluator in evaluation.}
\label{app-tab:evaluator}  
}
\end{table*}

\clearpage
\onecolumn
\section{Example Demonstration} \label{app-sec:example}
\noindent
\begin{tcolorbox}[enhanced,breakable,colback=white!95!gray,colframe=gray!50!black,rounded corners, title={A complete dialogue of User and ChatGE}]
\begin{lstlisting}[escapechar=|,breaklines=true,basicstyle=\small\ttfamily]
|\textbf{user:}|
[start interactive game development]

|\textbf{assistant:}|
<script>
</script>
<code>
</code>
<utter>
Hello, I am your game development assistant. Let's interactively develop a new poker game. Alright, let's start by configuring the essential parameters for our game. Could you provide the number of players and the betting range (minimum and maximum bet)?
</utter>

|\textbf{user:}|
There should be 3 players. The minimum bet should be 10 and the maximum bet should be 1000.

|\textbf{assistant:}|
<script>
script['Config']['Number of players'] = 3
script['Config']['Min bet'] = 10
script['Config']['Max bet'] = 1000
</script>
<code>
def config(self):
    self.n_players = 3
    self.min_bet = 10
    self.max_bet = 1000
</code>
<utter>
Great! We've set up the number of players and the betting range. Now, what about the suits and card value rank in the deck?
</utter>

|\textbf{user:}|
I want to use 4 suits: H, D, C, S. The card ranks should be from 2 to 12 and an additional rank 1.

|\textbf{assistant:}|
<script>
script['Config']['Suit'] = ['H', 'D', 'C', 'S']
script['Config']['Card value rank'] = [2, 3, 4, 5, 6, 7, 8, 9, 10, 11, 12, 1]
</script>
<code>
def config(self):
    self.n_players = 3
    self.min_bet = 10
    self.max_bet = 1000
    self.suit = ['H', 'D', 'C', 'S']
    self.value = [str(v) for v in [2, 3, 4, 5, 6, 7, 8, 9, 10, 11, 12, 1]]
</code>
<utter>
Perfect! The suits and card value rank have been configured. Do you need to rank the suits or should we ignore suit ranking?
</utter>

|\textbf{user:}|
No need to rank the suits.

|\textbf{assistant:}|
<script>
script['Config']['Suit have rank'] = False
</script>
<code>
def config(self):
    self.n_players = 3
    self.min_bet = 10
    self.max_bet = 1000
    self.suit = ['H', 'D', 'C', 'S']
    self.suit_have_rank = False
    self.value = [str(v) for v in [2, 3, 4, 5, 6, 7, 8, 9, 10, 11, 12, 1]]
</code>
<utter>
Got it! Suits won't have any inherent ranking. What about the card combinations rank? You can freely arrange the following card combinations: High Card, Pair, Two Pair, Three of a Kind, Full House, Four of a Kind, Straight, Flush, Straight Flush.
</utter>

|\textbf{user:}|
High Card, Pair, Two Pair, Three of a Kind, Straight, Flush, Full House, Four of a Kind, Straight Flush

|\textbf{assistant:}|
<script>
script['Config']['Card combinations rank'] = ['High Card', 'Pair', 'Two Pair', 'Three of a Kind', 'Straight', 'Flush', 'Full House', 'Four of a Kind', 'Straight Flush']
</script>
<code>
def config(self):
    self.n_players = 3
    self.min_bet = 10
    self.max_bet = 1000
    self.suit = ['H', 'D', 'C', 'S']
    self.suit_have_rank = False
    self.value = [str(v) for v in [2, 3, 4, 5, 6, 7, 8, 9, 10, 11, 12, 1]]
    self.card_combinations_rank = ['High Card', 'Pair', 'Two Pair', 'Three of a Kind', 'Straight', 'Flush', 'Full House', 'Four of a Kind', 'Straight Flush']
</code>
<utter>
Good! The configuration is complete. Now, let's move on to the phase. Please provide the Phase details.
</utter>

|\textbf{user:}|
The phase start: Configure the game and prepare the deck and chips for all players.

|\textbf{assistant:}|
<script>
script['Phase']['start'] = 'Configure the game, prepare the deck and chips for all players.'
</script>
<code>
def start(self):
    self.deck = []
    self.community = []
    for v in self.value:
        for s in self.suit:
            self.deck += [''.join([s,v])]
    
    for i in range(self.n_players):
        self.players[f'p{i+1}'].hole = []
        self.players[f'p{i+1}'].fold = False
</code>
<utter>
Great. The phase 'start' has been added. Are there any other phases?
</utter>

|\textbf{user:}|
Good. There is a phase called 'shuffle', which shuffles the deck.

|\textbf{assistant:}|
<script>
script['Phase']['shuffle'] = 'Shuffle the deck.'
</script>
<code>
def shuffle(self):
    shuffle(self.deck)
</code>
<utter>
Get! The phase 'shuffle' has been added. Are there any other phases?
</utter>

|\textbf{user:}|
I'd like to add a Phase 'blind' where we randomly select two players as the big blind and small blind. The small blind and big blind both place the minimum bet.

|\textbf{assistant:}|
<script>
script['Phase']['blind'] = 'Randomly choose two players as big blind and small blind respectively. Place minimum bet for small blind and minimum bet for big blind.'
</script>
<code>
def blind(self):
    def bet(player_id, amount):
        self.players[player_id].bet += amount
        self.players[player_id].remain -= amount
        
    small_blind, big_blind = random_choice(self.player_ids, 2)
    bet(small_blind, self.min_bet)
    bet(big_blind, self.min_bet)
</code>
<utter>
Alright, I've added the blind phase. Are there any other phases?
</utter>

|\textbf{user:}| 
There should be a Phase called 'dealx' that deals 'x' cards to each player.

|\textbf{assistant:}|
<script>
script['Phase']['dealx'] = 'Deal x cards to each player.'
</script>
<code>
def dealx(self, x):
    for i in range(x):
        for p in self.players:
            self.players[p].hole += [self.deck.pop()]
</code>
<utter>
OK. The phase 'dealx' has been added. Are there any other phases?
</utter>

|\textbf{user:}|
switch: Query each player to discard a number of cards and draw the same number of new cards from the deck.

|\textbf{assistant:}| 
<script>
script['Phase']['switch'] = 'Query each player to discard a number of cards and draw the same number of new cards from the deck.'
</script>
<code>
def switch(self, switch_indices, player_id):
    for i in switch_indices:
        self.players[player_id].hole[i] = self.deck.pop()
</code>
<utter>
Good! The phase 'dealx' has been added. Are there any other phases?
</utter>

|\textbf{user:}|
bet: Query each player to bet until all unfolded players have placed the highest bet or there is only one unfolded player.

|\textbf{assistant:}| 
<script>
script['Phase']['bet'] = 'Query each player to bet until all unfolded players have placed the highest bet or there is only one unfolded player.'
</script>
<code>
def bet_done(self, wait_to_bet):
    all_bet = [self.players[p].bet for p in self.get_unfold_players()]
    if not wait_to_bet and all([b==all_bet[0] for b in all_bet]):
        return True
    return False
</code>
<utter>
OK. The phase 'bet' has been added. Are there any other phases?
</utter>

|\textbf{user:}|
There should be a flopx phase, where discard one card from the deck and flop x cards from the deck to the community.

|\textbf{assistant:}| 
<script>
script['Phase']['flopx'] = 'Discard one card from the deck. Flop x cards from the deck to the community.'
</script>
<code>
def flopx(self, x):
    self.deck.pop()
    for i in range(x):
        self.community += [self.deck.pop()]
</code>
<utter>
OK. The phase 'flopx' has been added. Are there any other phases?
</utter>

|\textbf{user:}|
No, that's all phases for now.

|\textbf{assistant:}| 
<script>
</script>
<code>
</code>
<utter>
Great! The Phase is complete. Now, let's set up the flow for the game with these phases and the 'show' and 'prize' addtionally at the end.
</utter>

|\textbf{user:}| 
The flow should be: start, shuffle, blind, deal2, bet, flop3, bet, flop1, bet, flop1, bet, show, prize.

|\textbf{assistant:}| 
<script>
script['Flow'] = ['start', 'shuffle', 'blind', 'deal2', 'bet', 'flop3', 'bet', 'flop1', 'bet', 'flop1', 'bet', 'show', 'prize']
</script>
<code>
def set_flow(self):
    self.flow = ['start', 'shuffle', 'blind', 'deal2', 'bet', 'flop3', 'bet', 'flop1', 'bet', 'flop1', 'bet', 'show', 'prize']
</code>
<utter>
Excellent! The flow is set up, everything is ready. You have successfully created a new game!
</utter> 
\end{lstlisting}
\end{tcolorbox}

\end{document}